\documentclass[letterpaper, 10 pt, conference]{ieeeconf}  

\IEEEoverridecommandlockouts                              
\usepackage{graphicx} 
\usepackage{epsfig} 
\usepackage{mathptmx} 
\usepackage{times} 
\usepackage{amssymb}  
\usepackage{cite}
\usepackage{algorithm}
\usepackage[noend]{algpseudocode}

\title{\LARGE \bf
Augmented Reality for Human-Swarm Interaction in a Swarm-Robotic Chemistry Simulation
}

\author{Sumeet Batra, John Klingner and Nikolaus Correll
\thanks{S. Batra is supported by the University of Colorado Undergraduate Research Opportunities Program (UROP). }
\thanks{All authors are with the Department of Computer Science, University of Colorado at Boulder, Boulder, CO 80309, USA.}%
}

\begin{document}

\maketitle
\thispagestyle{empty}
\pagestyle{empty}

\begin{abstract}
We present a method to register individual members of a robotic swarm in an augmented reality display while showing relevant information about swarm dynamics to the user that would be otherwise hidden. Individual swarm members and clusters of the same group are identified by their color, and by blinking at a specific time interval that is distinct from the time interval at which their neighbors blink. We show that this problem is an instance of the graph coloring problem, which can be solved in a distributed manner in $O(log (n))$ time. We demonstrate our approach using a swarm chemistry simulation in which robots simulate individual atoms that form molecules following the rules of chemistry. Augmented reality is then used to display information about the internal state of individual swarm members as well as their topological relationship, corresponding to molecular bonds.  
\end{abstract}

\section{INTRODUCTION}

Swarm robotics bears great promise in applications that benefit from sampling the environment at high resolution and with adaptive density such as environmental monitoring, agriculture or surveillance \cite{csahin2004swarm,brambilla2013swarm} or as model system for natural swarming systems ranging from social insects to chemistry \cite{randall2016simulating}. Users usually interact with such systems via spatial representations of data the swarm is collecting, provide ensemble instructions, and let the swarm members decide on their individual actions in some distributed way \cite{bashyal2008human,naghsh2008analysis,vasile2011integrating,kolling2013human}. In the absence of spatial information, it is not clear how to communicate meaningful data and results of a swarm beyond simple lights, motion patterns \cite{podevijn2012self}, or statistical analysis of trajectories \cite{correll2006swistrack}. Thus, we propose combining our custom built swarm implementation with modern techniques from the emerging field of Augmented Reality (AR) to create an AR framework that effectively communicates relevant information about individual units of a swarm in a clear and visualizable manner, thus revealing otherwise hidden information about a swarm's behavior \cite{daily2003world,benavides2015invisibilia}. In addition, we introduce key challenges involved in developing scalable AR systems for scalable swarm implementations, and how our novel AR framework addresses those challenges. 

\section{RELATED WORK}

\begin{figure}[!htb]
\centering
\includegraphics[trim={25cm 0 25cm 1cm},clip,width=0.9\columnwidth]{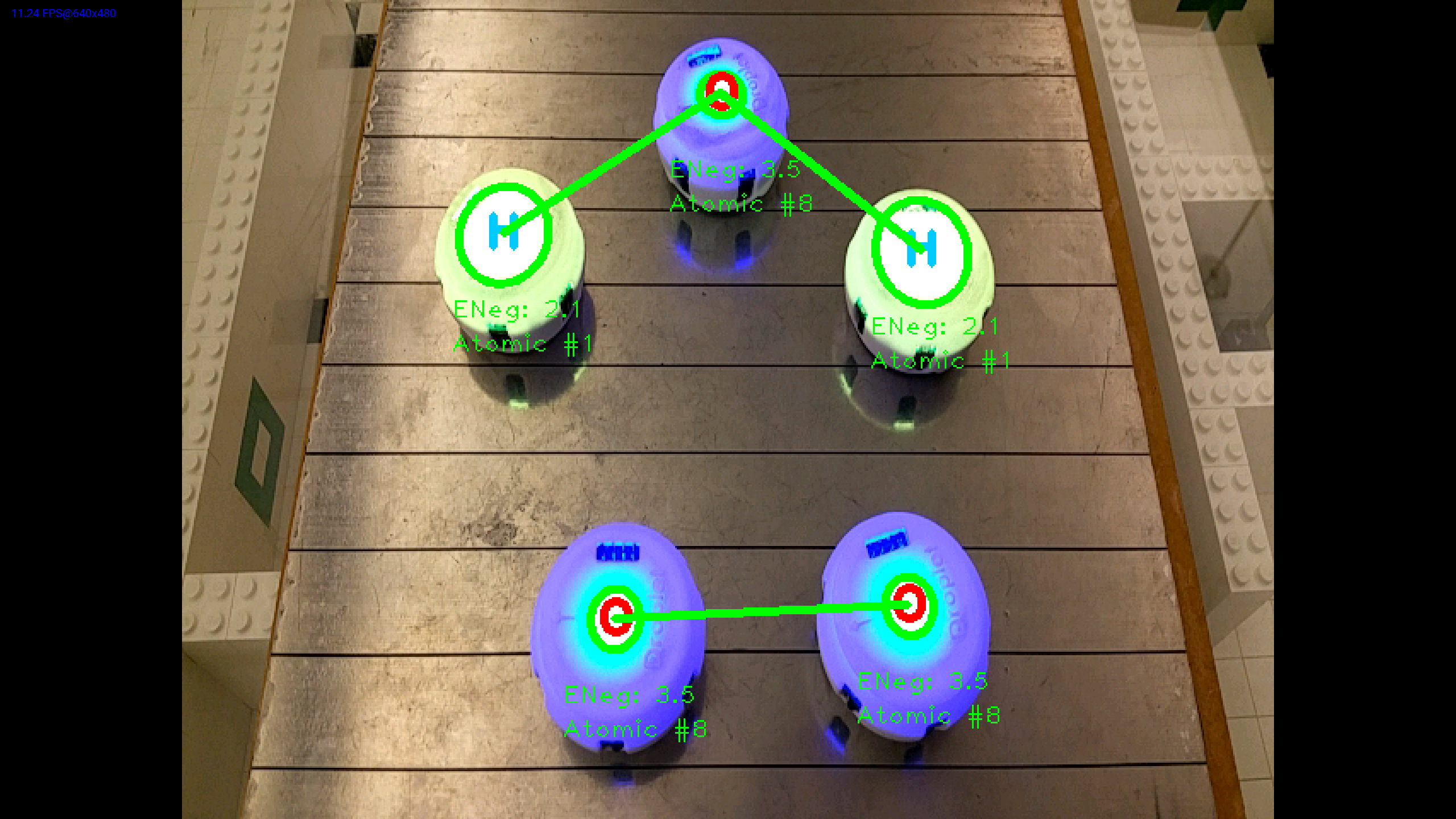}
\caption{Five droplets forming an $O_2$ and a $H_20$ molecule. Augmented Reality on a smart phone is used to show the actual bonds, electronegativity and atomic number. \label{fig:example} \vspace{-10px}}
\end{figure}

\subsection{Robot Swarm Implementation}

The basis of our swarm implementation are programmable ``Droplets'' --- small, round, cylindrical robots with locomotion and range-and-bearing sensors \cite{farrow2014miniature} for communicating and interacting with nearby Droplets that are engaged in simulating molecular self-assembly \cite{randall2016simulating}. Here, each robot takes the role of a single atom that assembles into complex molecules following the basic rules of chemistry using concepts such as free electrons and electro-negativity. Due to the complexity of the information involved, such as bonds between atoms or the Gibbs' free energy of a molecule, color change alone is not sufficient to communicate information to an user. In addition, information that is relevant for bonding changes with the configuration the simulated atom currently is in. 

\subsection{Relevant Chemistry}
This section briefly describes the chemical model that the Droplets implement as a tool to instruct high school students on chemistry. The reader is referred to \cite{randall2016simulating} for more details. 
    
We replicate Bohr's model for representing atoms and atomic interactions. Given it's simplicity, Bohr's model succeeds in communicating basic atomic structures and fundamental chemistry concepts. The central positively charged nucleus composed of protons and neutrons is surrounded by orbiting electrons at various energy levels. Free spots on the orbits determine which types of atoms can bond. This model falls short once we cross into the realm of modern quantum mechanics, however, and often leads to confusion as it is difficult to explain why atoms as simple as Hydrogen and Oxygen (Figure \ref{fig:example}) form a large variety of molecules not limited to $H_20$ (albeit at much lower likelihood). 


\begin{figure*}[htb!]
  \centering
  \includegraphics[height=0.17\textwidth]{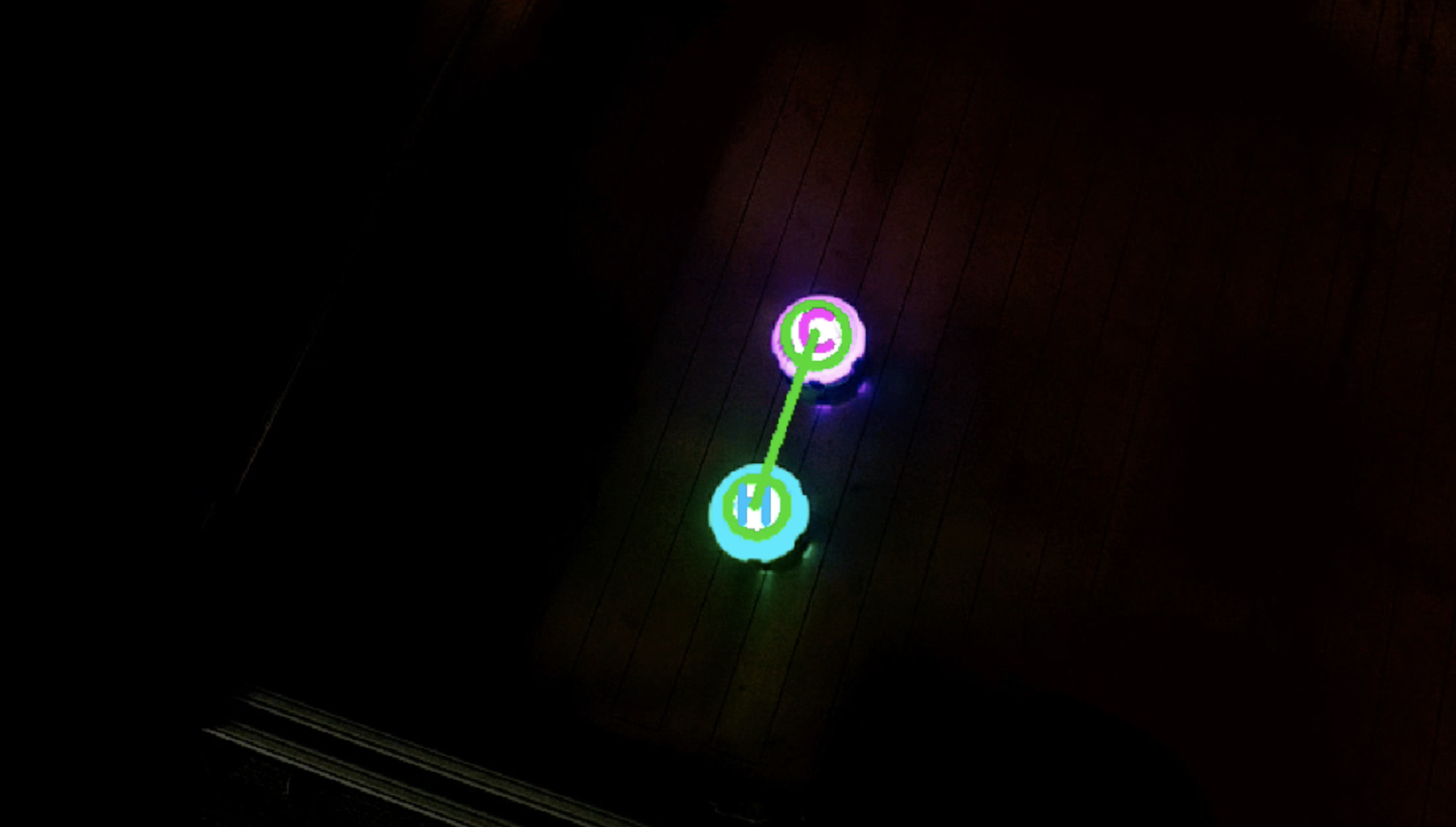}
  \includegraphics[height=0.17\textwidth]{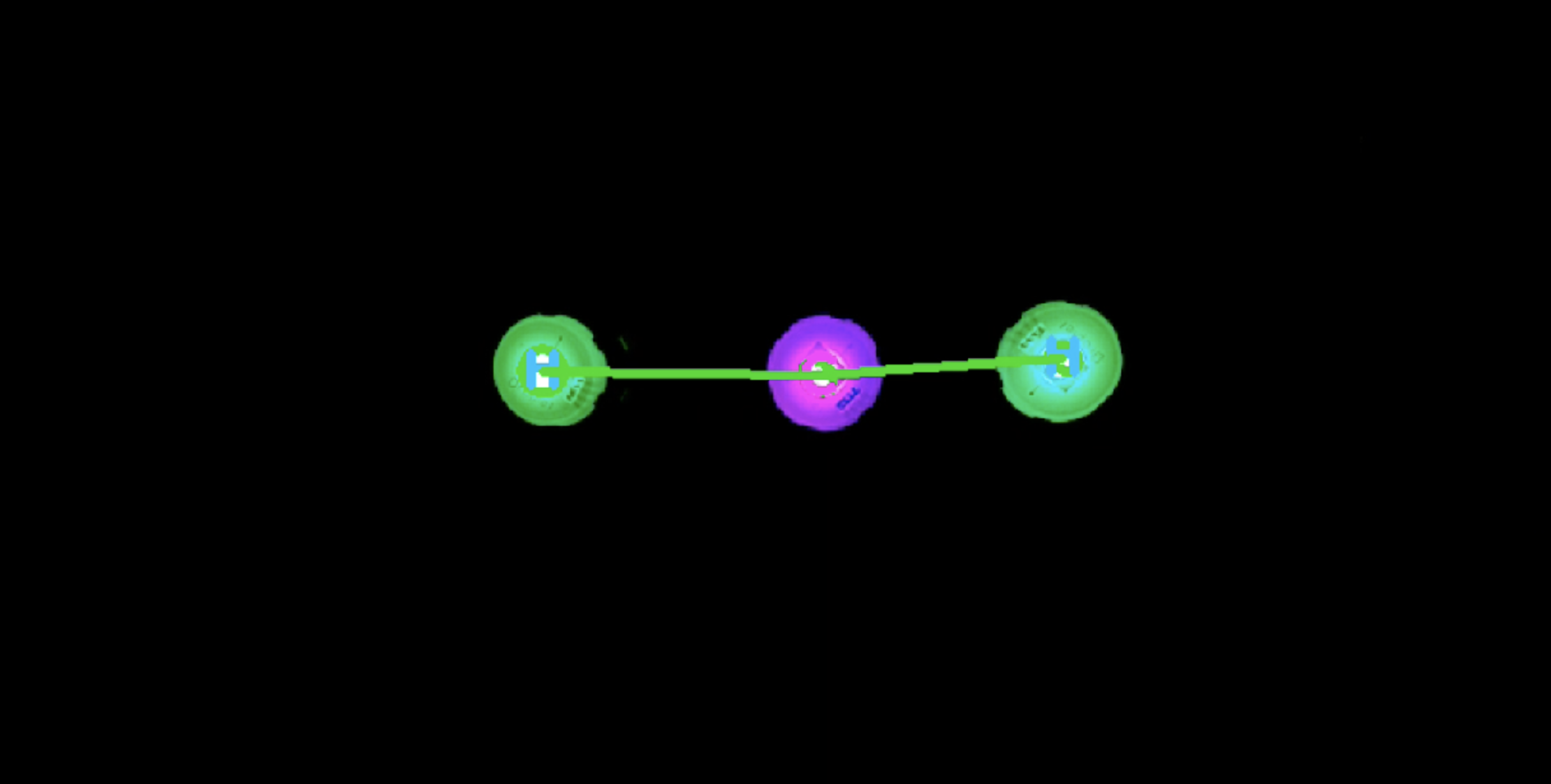}
  \includegraphics[height=0.17\textwidth]{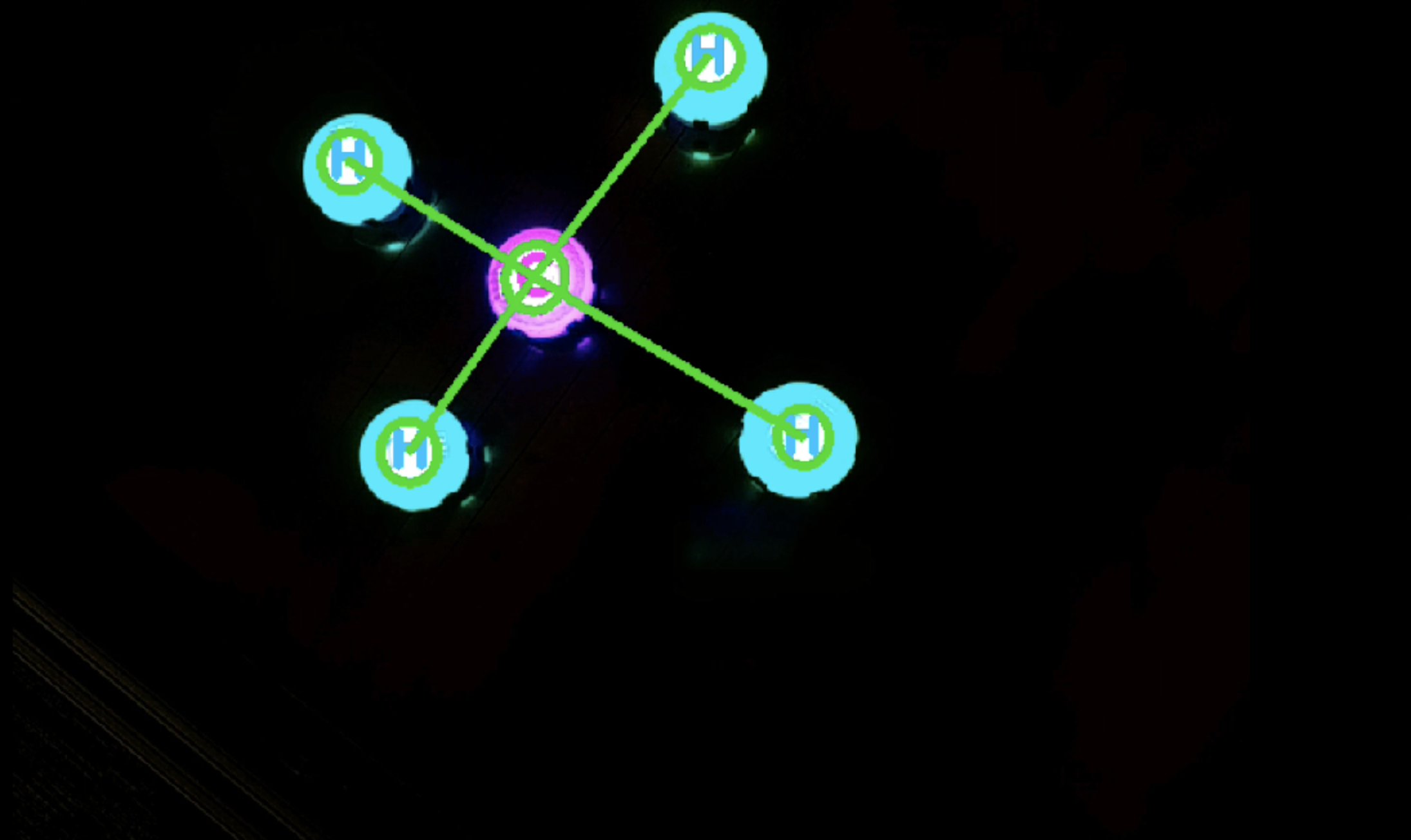}
    \caption{From left to right: First stage of $CH_4$ bonding. Carbon first bonds with a Hydrogen. Second stage of $CH_4$ bonding. Carbon bonds with an additional Hydrogen. Third stage of CH4 bonding. Carbon bonds with all 4 Hydrogens.}
\end{figure*}

Indeed, since Bohr's model was based solely on the hydrogen atom, it loses accuracy when we begin discussing larger atoms with many electrons at a wider range of energy levels. In addition, the model violates Heisenberg's Uncertainty Principle, which states that we cannot know the exact position and velocity simultaneously for a quantum particle. We therefore also employ information about each atom's electronegativity and the molecules Gibbs' free energy, making the robotic simulation much more complex than a stick-and-ball model, the go-to model for chemistry education.  


\subsection{Representing the Chemical Model with Droplets}

Each Droplet represents one atom using Bohr's model to communicate relevant data about itself and performs a random walk in the experimental arena. The atom represents its energy levels and orbiting electrons as an electronegativity value measured on the Pauling scale. We believe that electronegativity is the best measurement to communicate this data due to its importance in other quantum mechanics such as bond formation and molecular structures. An electronegativity value is assigned to each Droplet. Once Droplets are in proximity to each other, they exchange information and use their electronegativity values to determine whether bonds can form between, and if so, what the structure of the molecule is.  Here, the educational value is limited if the observer cannot differentiate individual molecules, need to remember what the different colors are, and look up statistical data for each atom, such as its electronegativity or atomic number. 

\subsection{Augmented Reality for Robotics}
Other works have proposed AR frameworks for tracking individual robots in swarm robotics applications \cite{reina2017ark} \cite{ghiringhelli2014interactive}. Augmented Reality for Kilobots (ARK), for example, implements a flexible AR interface for tracking swarms of Kilobot robots \cite{rubenstein2014kilobot}: small, inexpensive robots that can be used for swarm robotic applications. One downside of these proposed methods is that they rely on fixed, overhead cameras that introduce a degree of inflexibility in that the implementation is not easily transferable to different environments. Our AR framework has been implemented so that it may run in real-time on commercially available, low-cost cameras, such as those that come equipped on smartphones. This allows for greater mobility, allows for experiments and learning to occur in different environments, and makes our method very accessible. In addition, our swarm robotics platform is implemented in a way such that object persistence is possible, thus allowing our AR application to track and maintain unique information for each robot present in the experimental area. 

\section{Tangible Robotics and Augmented Reality for Human-Swarm Interaction}
Tangible interfaces have shown to be more successful in enhancing learning than digital systems \cite{marshall2007tangible}. Studies have shown that students who interact with tangible systems are more likely to be engaged in the learning process and that physical, three-dimensional objects are perceived and understood better than two-dimensional representations \cite{klemmer2006bodies}. Thus, we believe our swarm robotics platform is an excellent gateway for students to physically interact with molecular representations that would otherwise be unavailable to them. Combined with our novel AR framework, students will also be able to visualize the interactions occurring between atoms and molecules in real-time. 

\subsection{Augmented Reality Implementation}
We have developed a novel Augmented Reality framework that, with few adaptations, works with our swarm robotics platform that allows us to visually communicate otherwise complex, abstruse concepts regarding molecular interactions. The framework utilizes color information to determine element information and blinking patterns of groups of Droplets to determine molecular subgroups. Here, we describe the framework in more detail and address several challenges involved in creating a scalable AR framework for our scalable swarm robotics platform.

The key challenge in using augmented reality for human-swarm interaction is to efficiently communicate information to the augmented reality tool. In other words, how can information be encoded in a scalable way? In this paper, we propose to use synchronous blinking to indicate robots that are part of the same molecule and colors to indicate atom type. This requires neighboring molecules to coordinate on the blinking time interval, which is an instance of the distributed graph coloring problem \cite{kuhn2006complexity}. Specifically, no neighboring molecule can blink during the same time slots, which are the equivalent of colors in the graph coloring problem. As the bandwidth of color change and blinking is limited, some information still needs to be reconstructed by analyzing the local topology of the robots. 

We propose to use augmented reality to augment a display of the swarm in a cell phone display with additional information about the swarm. 

The augmented reality tool was developed using OpenCV4Android with OpenCV 3.1.0, an open source computer vision library for mobile application development. Due to the limitations of mobile computing and the demands of image processing, we decided that the best approach to utilizing OpenCV on an Android device was to write a majority of the application in C++ and have the device run the image processing natively. Using a Java Native Interface (JNI), the Java code can directly call OpenCV C++ functions as they are needed. This effectively allows the image processing functions to bypass the Java Virtual Machine (Dalvik VM) and run directly on the Linux Kernel. While this approach requires recompiling the native code for different hardware architectures, a significant performance boost is noticeable during runtime.  


Since the Droplets visually communicate their element type to users by color, we decided that a combination of color based tracking and contour detection was appropriate to identify, track, and correctly label each Droplet. Thus, on an abstract level, atomic information of each Droplet and molecular states of groups of Droplets are stored in their color and blinking pattern. 

The Droplets are first isolated in grayscale due to the simplicity of detecting high light intensity that each Droplet's LEDs produce. Proceeding a number of checks to verify a Droplet was found including contour shape and contour area, we discover the color of each Droplet by calculating each Droplet's average hue, saturation, and value. Each Droplet is then assigned to its respective element while its bond status and location are constantly updated.

If a subset of Droplets are bonded in a molecule, the molecular structure is determined by the preferred shape of the molecule in three dimensional space. This is primarily determined by the AR application, which takes into account the electronegativity assigned to each Droplet and the number of Droplets involved in the bonding process. In this application, this process is reduced to looking up a possible configuration for a group of atoms and the total Gibbs' free energy in a look-up table.

In addition, a central atom(s) is determined by the electronegativity of each atom within the molecule. For most cases, the central atom(s) is the element with the highest electronegativity; some special cases such as diatomic molecules are also recognized by the AR application thanks to the property that atoms within a diatomic have the same electronegativity, thus encompassing a wide range of practical molecular structures for this simulation. 

Bonded Droplets that blink red synchronously indicate to the AR interface that this subset of Droplets belong in a molecule. Using the electronegativity of each Droplet in the subset, the AR interface determines the orientation and bond pattern that each Droplet is making and displays this information to the user. In addition, a user can tap a Droplet on the display to toggle supplemental hidden information about each Droplet's atomic representation, including atomic mass, electronegativity, atomic number, etc.

Preventing neighboring molecules or other atoms from blinking at the same interval is another key challenge in this approach, and is addressed by reducing the problem to the distributed graph coloring problem and implementing a distributed implementation thereof. 

\section{SWARM COMMUNICATION (GRAPH COLORING)}

Droplets broadcast to the user that they are in a molecule by having every member of that molecule periodically flash red at the same time. However, we cannot guarantee that as the number of molecules present grows large, that each adjacent set of molecules flashes red at unique time intervals. Thus, it is important to prevent molecules with the same flashing period to be adjacent to each other. If we associate each unique time interval a molecule can flash red with a (virtual) color, then the problem is equivalent of not having any neighboring molecules with the same color. Assuming molecules are nodes of a graph and physically close corresponds to being edges between nodes, this problem is equivalent to the graph coloring problem. 

\begin{figure}[ht!]
\hbox{\hspace{-1.5em}\includegraphics[width=0.5\textwidth]{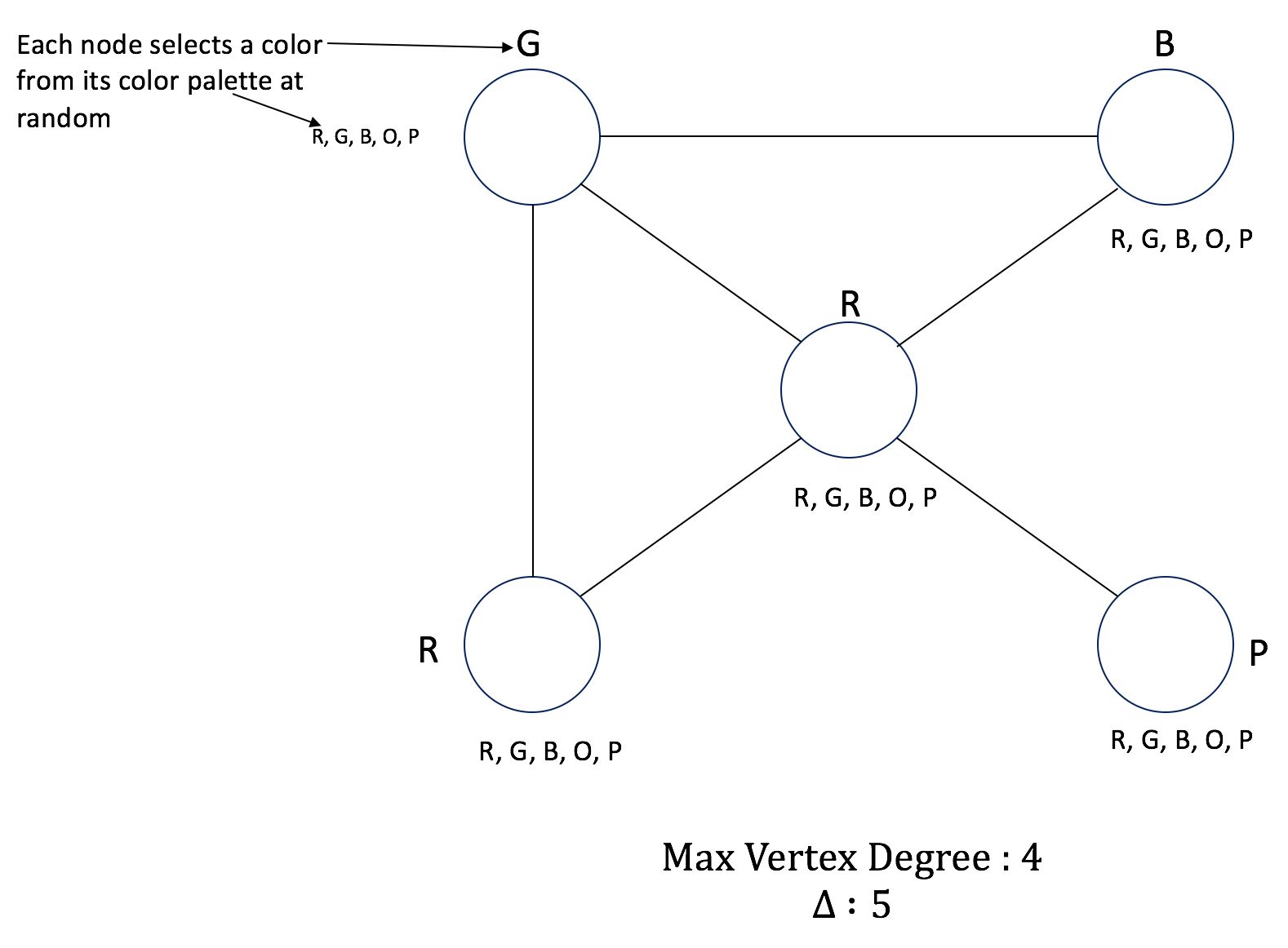}}
\hbox{\hspace{-1.0em}\includegraphics[width=0.5\textwidth]{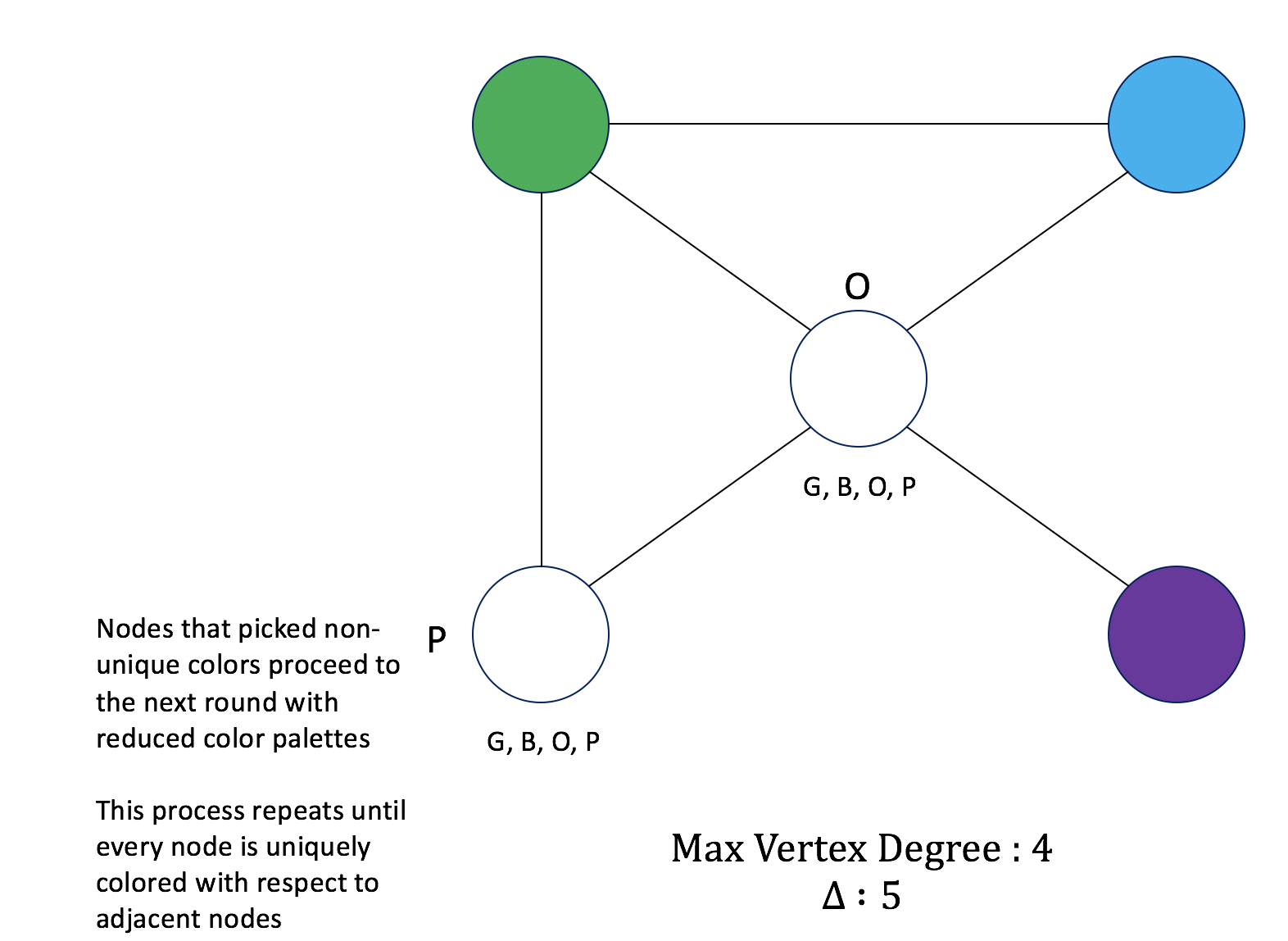}}
\caption{Example progression of the graph coloring algorithm. The top left, top right, and bottom right nodes chose unique colors relative to their neighbors, so their colors are fixed and they no longer participate. \label{fig:illustration}, \vspace{-10px}}
\end{figure}

The graph coloring problem itself is NP-Complete; however, in a distributed setting, the solution can be approximated \cite{johansson1999simple}. Let the center Droplet of each molecule represent a node in an undirected graph; an edge exists between two nodes if their respective center Droplets, and thus molecules, are adjacent. Given such a graph $G(V,E)$, and $\Delta + 1$ colors available to each node, where $\Delta$ is the highest vertex degree of the graph, the algorithm will complete in $O(log (n))$ rounds with high probability \cite{johansson1999simple}. A single round consists of two steps. First, each uncolored node randomly picks a color from its color palette, initially consisting of all $\Delta + 1$ colors. These nodes then compare their colors with their neighbors in the graph. If the chosen color is unique, the node is assigned that color and no longer participates. Otherwise, the node proceeds to the next round with its previously chosen color removed from its color palette. This process is illustrated in Figure \ref{fig:illustration} and the algorithm in pseudocode is provided in Algorithm 1.

\algblockdefx[Foreach]{Foreach}{EndForeach}[1]{\textbf{for each} #1 \textbf{do}}{\textbf{end for}}
\begin{algorithm}[ht!]
\caption{Distributed Graph Coloring Algorithm}\label{alg:euclid}
\begin{algorithmic}[1]
\Procedure{GraphColor}{}
\State $G\gets ErdosRenyiGraph(n, c)$ \Comment{Create random graph G with n vertices and average vertex degree c}
\State $\Delta \gets MaxVertexDegree(G, n)$
\Foreach{vertex $v$ \textbf{in} $G$}

\State $v.cp\gets ColorPalette(\Delta + 1)$

\EndForeach

\While{$numColored < n$}

\Foreach{uncolored vertex $v'$}

\State $v'.randColor \gets v'.cp.RandColor()$
\If {$\forall$ $v''$ \textbf{in} $v'.adjList$, $v''.randColor$ $\neq$ $v'.randColor$}

\State $v'.color \gets v'.randColor$ 
\State $numColored$++
\Else { $v'.cp.Remove(randColor)$} 
\EndIf
\EndForeach
\EndWhile
\EndProcedure
\end{algorithmic}
\end{algorithm}

We simulated this algorithm in C++ by first generating random graphs using the Erdos-Renyi model \cite{erdos1959random}. These graphs were then used to run the algorithm until all nodes in the graph were assigned a unique color relative to its neighbors. Under the assumption that each node is an independently functioning machine, thus allowing for parallel execution of the algorithm by each node, we found that the algorithm indeed completes in in $O(log (n))$ rounds with high probability. The simulation consisted of five iterations of creating random graphs from 100 nodes to 10,000 nodes with step sizes of 100 and running the prescribed algorithm on each graph.

\begin{figure}[h!]
\hbox{\hspace{0.0em}\includegraphics[width=0.5\textwidth]{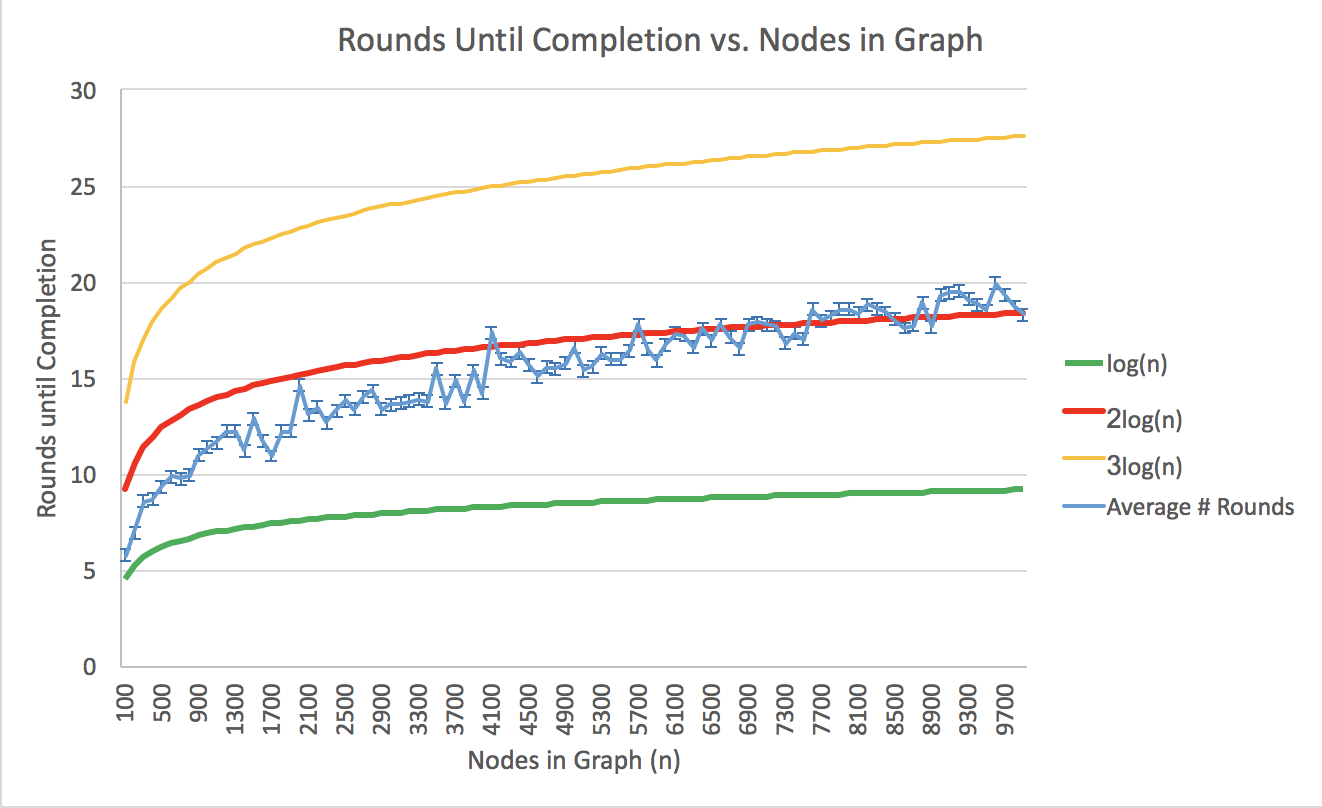}}
\caption{Average rounds until completion of the algorithm plotted with log(n), 2log(n), and 3log(n) curves for reference. The data was collected and averaged over 5 iterations of the algorithm, from graph sizes of 100 nodes up to 10,000 nodes, with increments of 100 and average vertex degree of 3  \label{fig:data}}
\end{figure}

All graphs had an average vertex degree of three. The data from the five iterations are averaged together and plotted with multiples of $log(n)$ in Figure \ref{fig:data}. It is clear from the data that the algorithm is $O(log(n))$ with a tight upper bound of $3log(n)$ and a tight lower bound of $log(n)$.

The simulation provides insight into the runtime behavior of the distributed graph coloring algorithm when applied to our swarm Droplet platform. For the purposes of our application, each molecule's center Droplet will choose a time interval to flash red. Adjacent center Droplets will compare their time intervals, and those molecules whose interval is unique will be assigned that interval. Those molecules who choose the same time interval must try again the next round. 

It is important to note that the number of unique time intervals is a fixed, physical constant, since too much overlap between two intervals will cause the AR's molecule recognition algorithm to confuse what should be two distinct molecules. Thus, the max vertex degree $\Delta$ of the network of molecules cannot exceed one less than the number of unique time slots in order for the AR application to function correctly.

We must also consider how real world factors, such as inter-Droplet communication, will affect performance. Specifically, if a Droplet fails to communicate some critical piece of information during a round, such as what color it chose for its group or which neighbor it is, then it must resend this information in the subsequent round, extending the overall runtime of the algorithm. Given that each Droplet has a, albeit low, probability of communication failure with adjacent Droplets, it is not unreasonable to assume that such events will occur in any trial or actual use of the program. Thus, the simulations provide a lower bound on the performance of the algorithm in an actual, physical setting. However, these faults should not affect the asymptotic behavior of the algorithm, and will instead add constant time overhead to the algorithm's performance.

\section{RESULTS AND DISCUSSION}

In order to understand how our application would perform in a real setting, we ran several experiments to test the robustness of the AR application under different circumstances. The variables we believe have the most impact on our AR application's Droplet detection accuracy are camera resolution, distance from the phone camera to the Droplets, the number of Droplets visible to the camera, and finally, the angle of inclination of the camera relative to the Droplets. Specifically, we address the following four questions in order to assess the robustness of our AR implementation: (1) To what degree does camera resolution affect performance of the application and Droplet detection accuracy? (2) How does Droplet detection accuracy change as a function of distance (3) How does the Droplet detection accuracy of our image processing pipeline change with respect to the number of Droplets visible in the camera frame? (4) How does Droplet detection accuracy change as a function of camera angle inclination relative to the horizontal? The experiments and results used to address these questions are detailed in the proceeding section. In addition, we provide empirical data that will be used to discuss the performance of Algorithm 1 in the real world. These results help us parameterize what set of variables allow the AR application to maintain a high degree of detection accuracy under different environments while also maintaining an acceptable framerate.

\subsection{Camera Resolution}

We empirically determined that the best resolution for the camera is 640x480 pixels. This resolution is a compromise between performance and quality of the video feed. Any lower resolution, while boosting performance, would result in a blurry and choppy image stream, and any higher resolution will significantly impact the performance of the application to default to non-ideal frame rates.
 Due to the fact that a Droplet's color is interpreted as an average hue, saturation, and value over a region of pixels, changing the pixel density can throw off these calculations, resulting in erroneous behavior and occasionally false positives in the application. Thus, we have decided to lock the resolution to 640x480 for all devices that run the application for the purpose of standardization.

\subsection{Accuracy as a Function of Distance}
The AR tool primarily relies on color thresholding in order to detect Droplets and determine their position relative to other Droplets, as well as compute their bond status. The light from the Droplets is interpreted as a high saturation and value in the HSV color space, thus making it easily distinguishable from background light and surrounding objects that are not Droplets. However, there is a certain threshold corresponding to the distance from the camera to the Droplets’ power board in which the distinction can no longer be made. This variable significantly contributes to the accuracy at which the application can detect Droplets, thus we found it useful to gather data on this subject. 

\begin{figure}[h!]
    \includegraphics[height=0.3\textwidth]{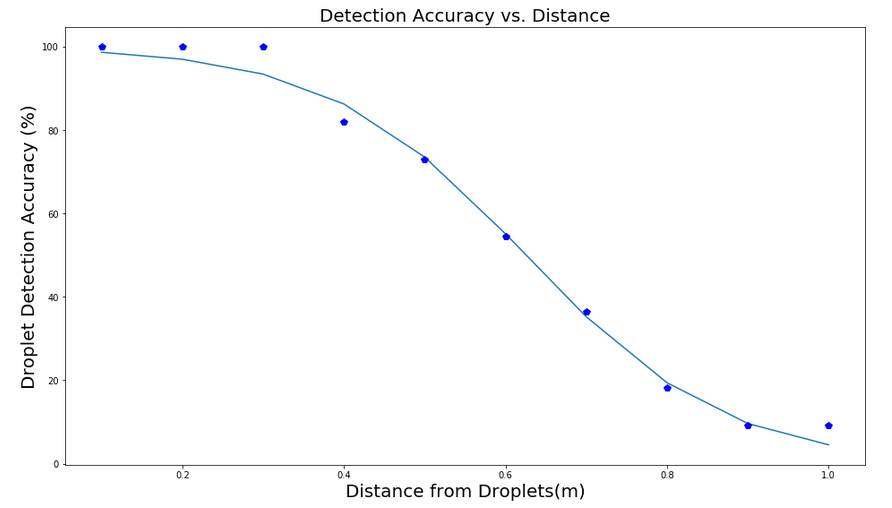}
    	\caption{Droplet detection accuracy as a function of distance. The sigmoid function was fitted to the data.\label{fig:angle}}
\end{figure}

Starting at an inclination of 0.1 meters relative to the floor until a height of 1 meters, with increments of 0.1 meters at a time, we measured the accuracy of the AR application as the ratio of correctly identified Droplets to total Droplets visible on the camera. At each increment, we kept a fixed camera angle of 0 degrees relative to the horizontal, or a birds eye view, and a fixed x-y position in each z-plane. The results are graphed in Figure \ref{fig:angle}.

The data shows that closer distances have fairly high accurate Droplet detection rates. However, beyond 0.6 meters, we begin to see nominal performance from the application. This affects the number of Droplets we can have at any given time while maintaining a high detection accuracy, since increasing the number of Droplets implies moving the camera further back in order to capture all the Droplets present. 

\subsection{Number of Droplets versus Detection Accuracy}

\begin{figure}[h!]
\includegraphics[width=0.45\textwidth]{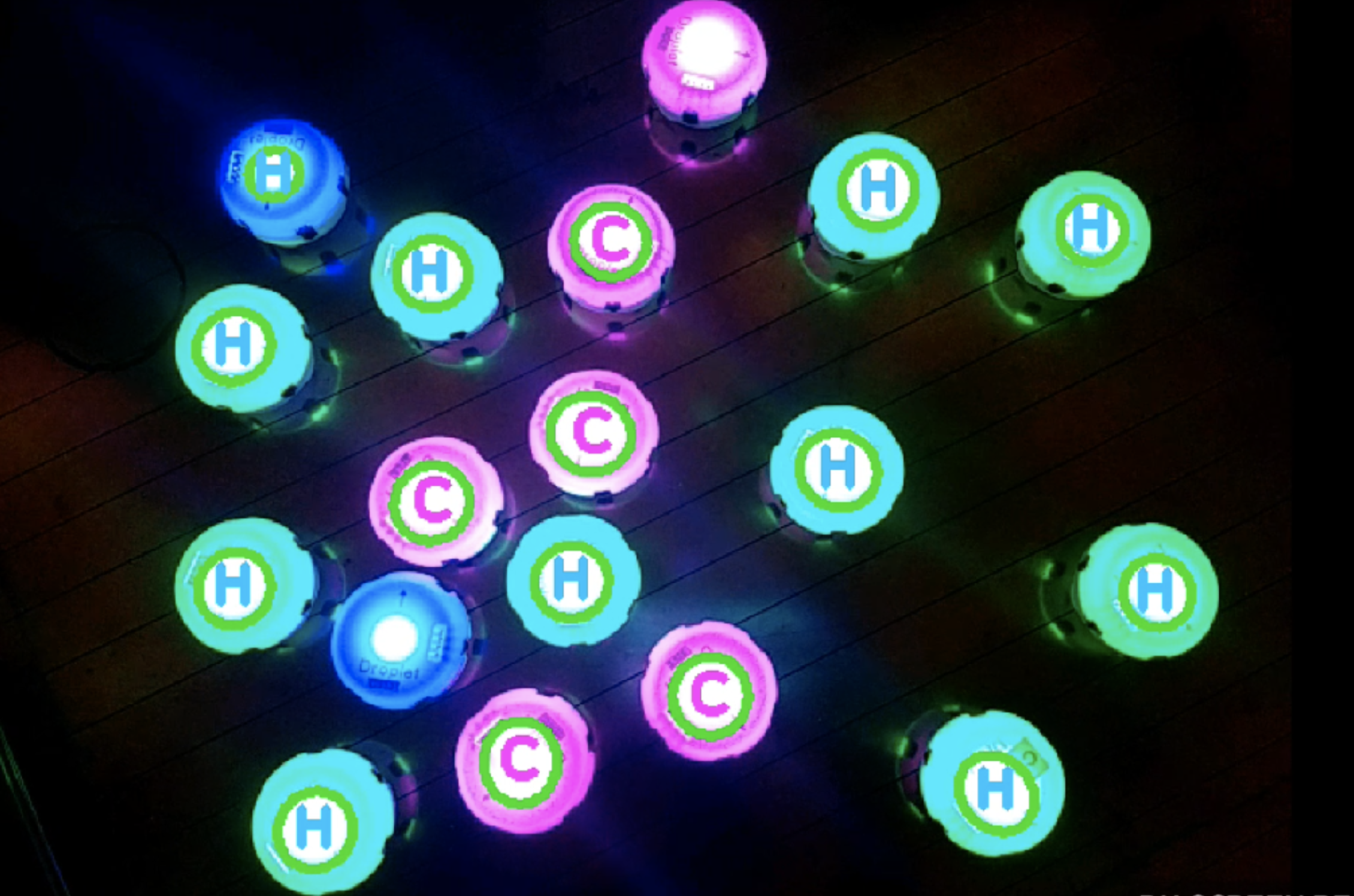}
\caption{Example of 18 Droplets, representing oxygen (blue), hydrogen (green) and carbon (pink). One oxygen is misclassified as hydrogen, and one oxygen and carbon are not identified.
\label{fig:max}}
\end{figure}

The maximum number of Droplets that the camera can detect at any instance in time is closely intertwined with the maximum distance of the camera to the power board. This is due to the fact that a greater distance between the camera and Droplets results in more Droplets being able to fit within the camera’s boundaries. However, another determining factor for the maximum allowed Droplets on the camera is the ability to detect individual Droplets and accurately represent them as their corresponding elements. As more Droplets are densely packed in a confined space, the color information from each Droplets becomes is lost and the resolution for each individual Droplet becomes too small. For example, Figure \ref{fig:max} shows an ensemble of 18 Droplets with the blue Droplets (Oxygen) either not or incorrectly classified, and one pink Droplet (Carbon) not identified. 

Thus, we believe that the number of Droplets on the field is another important variable to determining the robustness of our application. We performed experiments in order to understand the correlation between detection accuracy and the number of Droplets visible to the camera. We collected data on the detection accuracy from 1 to 20 Droplets at increments of 1 Droplet at a time. Once again, the camera angle of inclination was fixed at 0 degrees relative to the horizontal. For each increment, the detection accuracy was averaged over three trials. The results are graphed in the figure below.

\begin{figure}[h!]
	\includegraphics[height=0.3\textwidth]{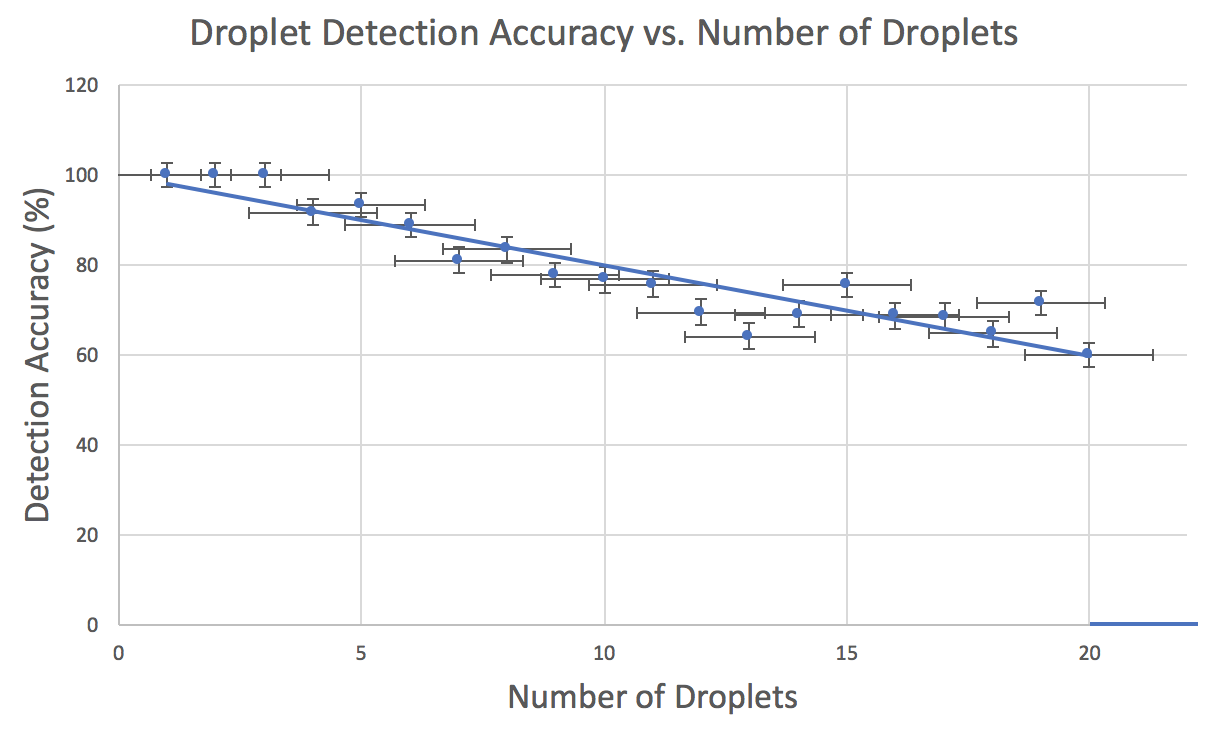}
    \caption{Droplet detection accuracy vs. number of Droplets visible to the camera. An inverse linear relation exists until the empirically determined max of 20 Droplets}
\end{figure}

We see an inverse linear relationship between the number of Droplets present and the Droplet detection accuracy up until 20 Droplets are present on the field, which we have empirically determined to be the maximum number of Droplets can detect at any given time. At 20 Droplets, individual Droplet color information is lost, and light over-saturation prevents the camera from perceiving any of the Droplets, as we can see from the stark drop of detection accuracy to 0 percent at 21 Droplets. 

\subsection{Angle of Inclination}

The final variable in question is how the angle of inclination of the camera affects the performance of our AR application. In practice, as the angle of inclination increases, the surface area of colored light emitted from the Droplets and perceived from the camera reduces until the algorithm no longer considers the light source as a Droplet candidate. How this affects detection performance at different camera angles is of interest and importance, as we do not expect the average user to hold their smart phones at a perfect birds-eye-view indefinitely. 

The experiment was performed starting at 0 degrees relative to the horizontal to 90 degrees, at increments of 15 degrees at a time. The distance from the Droplets was kept fixed at 0.3 meters. In order to de-correlate angle of inclination and number of Droplets present, we repeated the experiment at each increment with 5, 8, and 11 Droplets, and averaged the results. The results are shown in the figure below.

\begin{figure}[h!]
\includegraphics[height=0.3\textwidth]{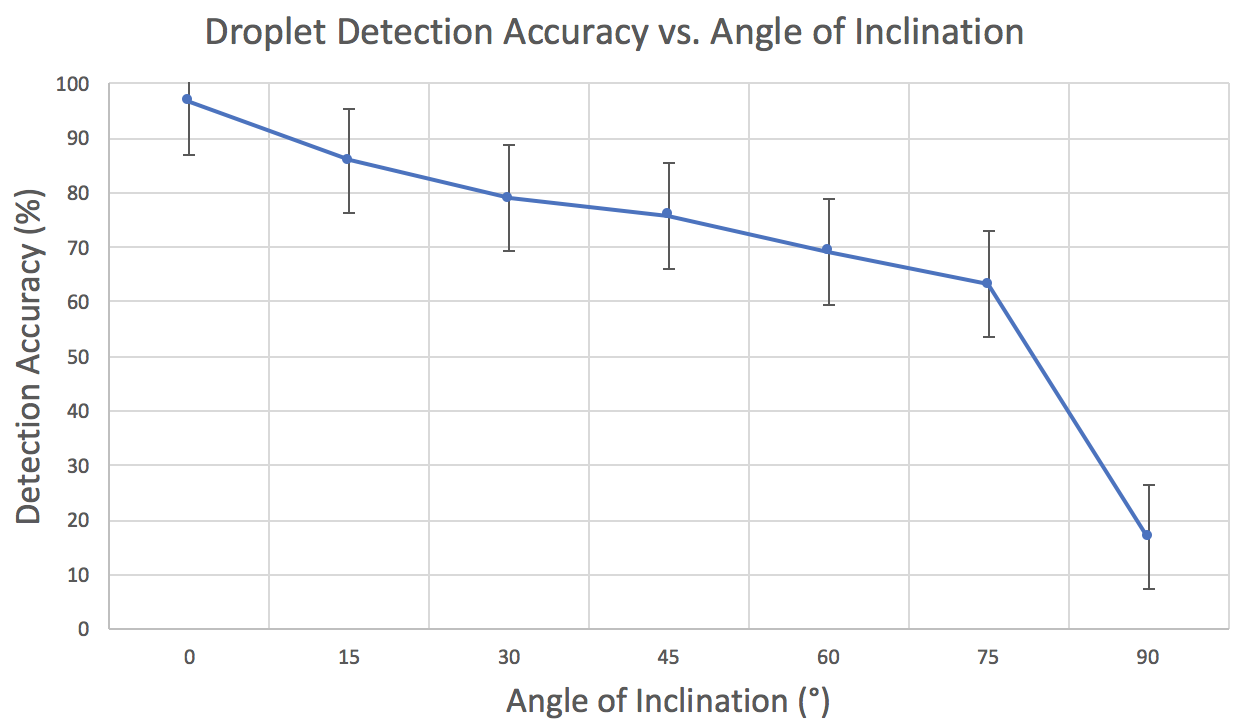}
	\caption{Droplet detection accuracy vs. angle of inclination. The detection accuracy is fairly high until the angle steepens to 90 degrees}
\end{figure}

While angle of inclination and detection performance are negatively correlated, camera angle inclination does not affect performance as severely as number of Droplets present or camera distance for angles between 0 and 75 degrees. For steeper angles higher than 75 degrees relative to the normal, we do notice a significant drop in detection performance, for the reasons listed above. 

\subsection{Performance of Algorithm 1 on Real Data}
Algorithm 1 entails our approach for handling molecule formation of different groups of Droplets using each group's respective red blinking pattern. We drew from the Graph Coloring Problem to formulate our algorithm, and showed that the algorithm converges and leads to successful formation of all distinct molecules in $O(log(n))$ time. 

\begin{figure}[h!]
\includegraphics[height=0.28\textwidth]{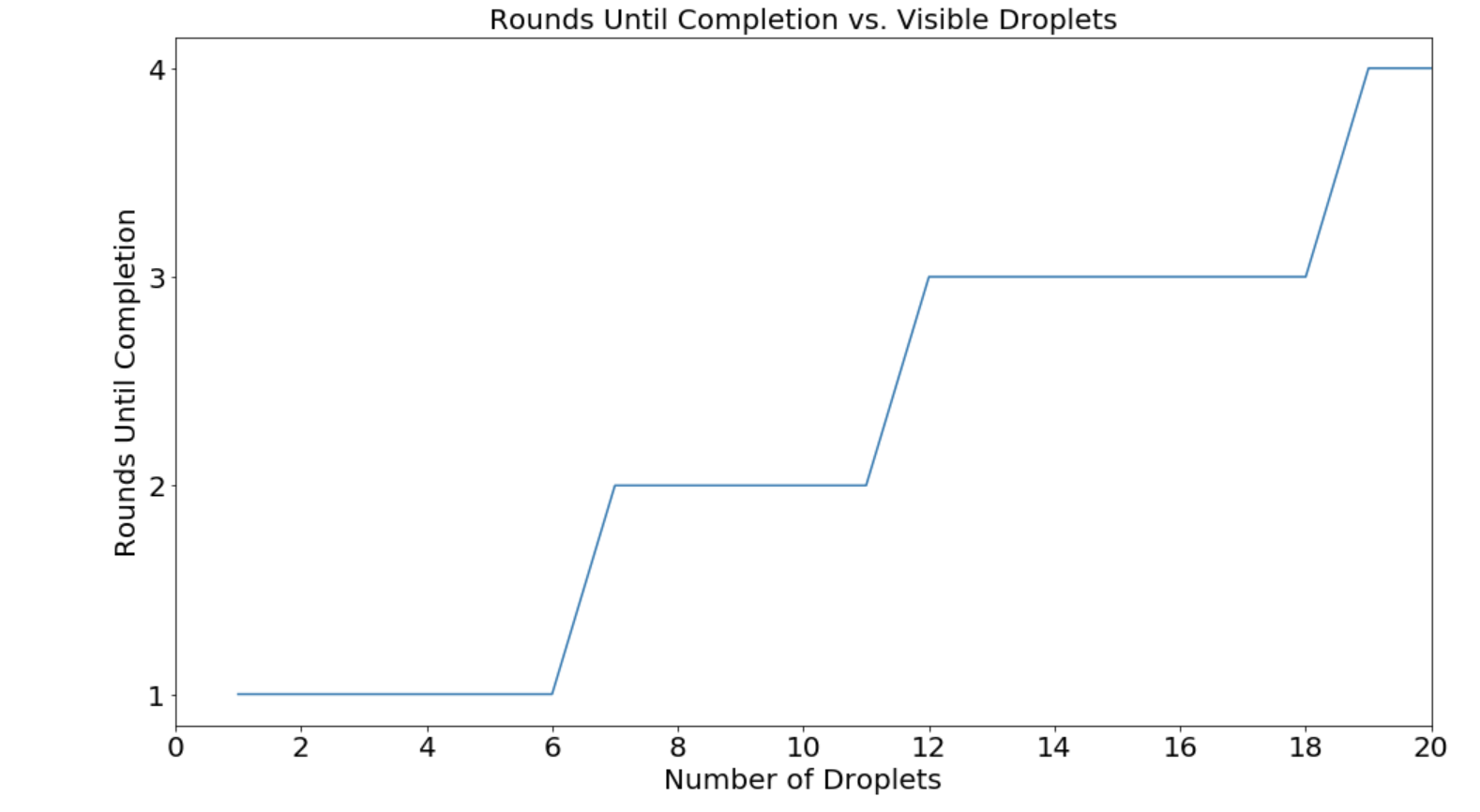}
	\caption{Rounds of red blinking needed until molecule assignment completed by the AR app. \label{figure:red}}
\end{figure}

To analyze the performance of this algorithm in a realistic scenario, we introduced additional Droplets one at a time to different molecule groups and measured how many rounds of red blinking were required before our AR application could recognize the correct number of molecules present in the experimental arena. The results are graphed in Figure \ref{figure:red}.

\newpage

From the data above, we can see that the algorithm behaves in accordance with the results found in the simulation we ran for Algorithm 1 when the number of Droplets in the experimental area is relatively low. However, we begin to see three to four rounds of red blinking required as the number of visible Droplets approaches 20. We empirically found that our image processing pipeline made several mistakes assigning Droplets to their correct groups as we increased the number of Droplets present. We believe this to be a result of the camera intrinsics, specifically resolution and framerate, that causes certain Droplets that blink several hundred milliseconds apart to appear to blink at roughly the same time. Thus, we conclude that while our results from the simulations run with Algorithm 1 give us a good theoretical upper bound on performance, noisy data in real world settings causes a reduction in performance of our algorithm.




\section{CONCLUSION}

In an environment where swarm members are equipped with GPS information, such as in an environmental monitoring or agricultural task, the proposed approach can still be used to differentiate between multiple robots that are visible at the same time, for example swarming aircraft that flock in close proximity. In future work, we wish to use this information to explicitly address individual swarm members and provide control signals. In the chemistry simulation, this could be steering individual atoms or breaking up molecules. In an environmental monitoring or agricultural application, this could be selecting a leader or other mission-specific commands. 

Based on the robustness and scalability of the Augmented Reality interface in conjunction with a properly implemented swarm robotics platform, we believe that this tool serves as the foundation for future swarm robotics research and development. In the chemistry simulation, we found the AR interface to be a practical tool for augmenting the information users could retrieve, which might be important in a teaching context. Having the Droplets maintain their respective molecular structures while the AR interface overlays the physical space with relevant atomic information effectively combines traditional 3D molecular models and two-dimensional molecular charts into one interactive tool. 

The augmented reality interface’s ability to visualize hidden data might have great implications for swarm robotics and robot automation research in general. Data aggregation can lead to faster development cycles for improved swarm behavior algorithms and automated robot behavior. This can have long-standing effects on modular robotics for education and/or swarm research, as well as on swarm nanorobotics for medical treatment, construction, etc. Clearly, swarm data aggregation and visualization can advance our understanding and utilization of swarms across multiple industries.
 
As previously discussed, the AR interface and the swarm implementation is not without limitations. However, many of these limitations are a byproduct of current computer limitations. Computer vision is processor intensive; only recently have processors reached a stage where Augmented and Virtual reality devices have become viable. As computer architectures continue to evolve, current limitations will continue to relax and Augmented Reality solutions will become more powerful and relevant.

\bibliography{bibfile} 
\bibliographystyle{ieeetr}

\end{document}